\pdfoutput=1

\documentclass[11pt]{article}

\usepackage[]{acl}

\usepackage{times}
\usepackage{latexsym}
\usepackage{booktabs}
\usepackage{multicol}
\usepackage{tabularx}
\usepackage{graphics}
\usepackage{soul}
\usepackage{blindtext}
\usepackage{microtype}
\usepackage{paralist}
\usepackage{diagbox}
\usepackage{multirow}
\usepackage{comment}
\usepackage{booktabs}
\usepackage{relsize}
\usepackage{array}
\usepackage{multirow}
\usepackage{graphicx}
\usepackage{xspace}
\usepackage{cleveref}
\usepackage[T1]{fontenc}

\usepackage[utf8]{inputenc}

\newcommand{\ex}[1]{\textit{#1}\xspace}

\newcommand{\token}[1]{\textbf{\textit{#1}}\xspace}
\newcolumntype{P}[1]{>{\centering\arraybackslash}p{#1}}

\title{LED down the rabbit hole: exploring the potential of global attention for biomedical multi-document summarisation}

\author{Yulia Otmakhova$^{1,*}$, Hung Thinh Truong$^{1,*}$, Timothy Baldwin$^{1,3}$,\\
\textbf{Trevor Cohn$^1$, Karin Verspoor$^{2,1}$, Jey Han Lau}$^1$\\
$^1$The University of Melbourne, $^2$RMIT University, $^3$MBZUAI\\
\smaller \texttt{\{yotmakhova,hungthinht\}@student.unimelb.edu.au},  \texttt{tb@ldwin.net}, \\
\smaller \texttt{trevor.cohn@unimelb.edu.au},
\texttt{karin.verspoor@rmit.edu.au},
\texttt{jeyhan.lau@gmail.com}}

\begin{document}
\maketitle

\begingroup\def\thefootnote{*}\footnotetext{Equal contribution}\endgroup

\begin{abstract}

In this paper we report on our submission to the Multidocument Summarisation for Literature Review (MSLR) shared task. Specifically, we adapt PRIMERA \citep{xiao-etal-2022-primera} to the biomedical domain by placing global attention on important biomedical entities in several ways. We analyse the outputs of the 23 resulting models, and report patterns in the results related to the presence of additional global attention, number of training steps, and the input configuration.

\end{abstract}

\section{Introduction}


In this paper we describe our experiments and results on the Multidocument Summarisation for Literature Review (MSLR) shared task.\footnote{\url{https://github.com/allenai/mslr-shared-task}} 
In particular, we attempt to improve on previous multi-document summarisation models in the biomedical domain, which have tried to integrate domain knowledge by marking important biomedical entities \citep{wallace2021generating,deyoung-etal-2021-ms}. We hypothesise that highlighting such entities by placing global attention on them will enable better aggregation and normalisation of related entities across documents, and thus improve the factuality of the generated summaries. To explore this idea, we experiment with four different ways of modifying the global attention mechanism of PRIMERA \citep{xiao-etal-2022-primera}, a recent state-of-the-art model designed for multi-document summarisation (MDS). In particular, while by default the global attention tokens in Primera are used to separate documents in the input and capture their relationships, we assign global attention to important biomedical entities in input documents to create links between them. Moreover, to examine the effect of content selection on the quality of summaries produced by this underlying model, we compare results where we use the whole abstract as input vs.\ only the concluding sentences (which we expect to be more informative). We train and analyse models in zero-shot, few-shot (10 and 100), as well as fully fine-tuned scenarios. Overall we evaluate (using both automatic metrics and human evaluation) a total of 23 models, two of which formed our official submissions to the leaderboard.\footnote{Additional results and code for all models is provided at \url{https://github.com/joey234/PRIMER-pico-attn}.} Both submitted models substantially outperform the baseline approaches \citep{deyoung-etal-2021-ms} in terms of automatic metrics, and one achieves the best performance in terms of BERTScore and ROUGE-2 among all submissions. Overall, our contributions in comparison to the previously published domain-specific models for MDS are the following: 
\begin{itemize}
    \item We explore the potential of using global attention as a means to highlight important biomedical entities, in order to improve aggregation across input documents.
    \item We examine how the amount of training data influences the quality of generated summaries, and propose several scenarios where the performance of few-shot and even zero-shot models is on par with that of fully fine-tuned ones.
    \item We show that in the fine-tuned scenario, the model is able to select important content without additional marking.
\end{itemize}

\section{Dataset}

We use the Cochrane dataset as provided in the shared task without any additional data. See \Cref{tab:cochrane_statistic} in Appendix \ref{dataset_stats} for dataset statistics.

\subsection{Pre-processing}

As the trials are collected automatically from the Cochrane library, they contain redundant metadata such as hyperlinks, 
trial identifiers, 
funding information, 
copyright statements, 
and publication records. 
We perform string matching using regular expressions to remove this content.
Following \citet{wallace2021generating}, for each review, we concatenate all corresponding documents and add a separator token to denote the end of each document. 

\subsection{Entity marking}

The PICO framework describes several essential components of the central question in a clinical trial, including Populations (e.g.\ \textit{diabetics}), Interventions (e.g.\ \textit{animal insulin}), Comparators (e.g.\ \textit{human insulin}), and Outcomes (e.g.\ \textit{glycaemic control}) \citep{huang2006evaluation}.
We tag PICO spans in input and target documents to make the summarisation models explicitly attend to them.
We train a tagger on the EBM-NLP dataset \citep{nye-etal-2018-corpus}, which contains annotations for the P, I, and O classes\footnote{Comparators are grouped with Interventions in the dataset due to the difficulty in distinguishing them.}
on abstracts of randomized controlled trials.
Using this dataset, we fine-tune the BioLinkBERT model \citep{yasunaga-etal-2022-linkbert}, a BERT variant that leverages links between documents that achieve state-of-the-art results on various biomedical NLP tasks, including the PI(C)O tagging task.
We adopt the same hyperparameters as in \citet{yasunaga-etal-2022-linkbert} using the BioLinkBERT\textsubscript{base} model, and achieve 74.06 macro-$F_1$ score on the EBM-NLP test set, which is comparable to the reported results in \citet{yasunaga-etal-2022-linkbert}.
We run the trained PIO tagger on the Cochrane dataset for both the documents and summaries. 
For simplicity, we only use two new special tokens \token{<ent>} and \token{</ent>} to mark the beginning and the end of each PICO span (e.g.\ \textit{\token{<ent>} Magnesium sulfate \token{</ent>} does not have a major impact on disease progression in \token{<ent>} women with mild preeclampsia \token{</ent>}.}).

\Cref{tab:cochrane_statistic} presents basic statistics of the Cochrane dataset used in this challenge. 
The average number of PIO spans in the summary and input documents is based on the output of the trained PIO tagger.
Note that target summaries for the test set are not provided to participants.

\section{Evaluation}

For the automatic evaluation, in addition to ROUGE scores \citep{lin-2004-rouge} and BERTScore\footnote{Hash code: \texttt{roberta-large\_L17\_no-idf\_ \\ version=0.3.11(hug\_trans=3.1.0)}} \citep{zhang2019bertscore}, we report the metrics introduced in \citet{deyoung-etal-2021-ms}, namely $\Delta$EI which measures the distance in predicted direction of the conclusions (\textit{increases}, \textit{decreases}, or \textit{no change}) in the target and generated summaries. For this metric, we report the average distance across samples and also macro-F1 score, in which the predicted direction for the target summary is treated as the correct label ($\Delta$EI-$F_1$).

To estimate quality of the generated summaries, especially in terms of their factuality, we also perform human evaluation, for which we adopt the binary decision method proposed in \citet{otmakhova-etal-2022-patient}. As we need to assess results from a large number of models, we simplify the evaluation, focusing only on factual errors and collapsing the categories of \textit{modality} and \textit{polarity} into a single category with five potential values (\textit{positive}, \textit{negative}, \textit{no effect}, \textit{no evidence}, \textit{no claim}), similar to how it was done by \citet{deyoung-etal-2021-ms}. Thus, we report if \textbf{PICO} elements used in the correct and generated summaries are aligned, if the \textbf{direction} of the findings is the same, and if the summaries are \textbf{factual}, that is, correct in these two aspects. In addition, to analyse common errors, we annotate generations as \textbf{contradictory} (i.e.\ containing statements with the same set of PICO elements but different polarity), \textbf{malformed} (i.e.\ including lexical and grammatical errors or repetitions), and \textbf{not evidential} (i.e.\ claiming that there is not enough evidence to determine the effect of intervention). We list some examples of contradictory, malformed and non-evidential summaries in Appendix \ref{sec:examples}.

As the vast majority of the target summaries were multi-aspect --- that is, contained statements regarding several groups of patients, interventions or outcomes --- one of the difficulties we experienced during the evaluation was comparing them to generated summaries which were either single-aspect or contained different sets of PICO elements. We adopted a precision-based approach when evaluating such pairs of summaries: while it is not necessary for the generated summary to contain all PICO elements included in the target to be considered correct, it must not include any extra PICO elements. In the case of extra PICO elements in the generated summaries, we compared them against the \textit{Objectives} section of the review's abstract to determine if they were truly erroneous or if the target conclusion underreported some of the elements. Moreover, in the case of multi-aspect summaries we consider direction to be correct only if it is correct for the corresponding set of PICO elements.

Thus though our evaluation approach is less detailed than the one proposed in \citet{otmakhova-etal-2022-patient}, it is more strict in terms of alignment of multi-aspect summaries.

\section{Experiments}

\subsection{Model}


We base our experiments on PRIMERA \citep{xiao-etal-2022-primera}, which was designed for multi-document summarisation, and 
experiment with zero-, 10-, 100-shot, and fine-tuning scenarios with the same hyperparameters reported by the authors of the paper. We use the same random seed for all models to ensure consistency. For the baseline model (\textit{No entity}) we use documents and summaries without any entity marking; all other models use documents with entity tags.

\begin{table}[t!]
\footnotesize
    \centering
    \begin{tabular}{p{2cm} p{5cm}}
    \toprule
    Setting & Description \\
    \midrule
    DocSep & The global attention is only set on the document separation token (\token{<doc-sep>}) as in the original PRIMERA model. The attention on \token{<doc-sep>} is used across the board in all settings described below. \\
    EntMarkers & In addition to the \token{<doc-sep>} global attention, we set global attention on tokens which mark the beginning and end of entities (i.e.\ \token{<ent>}, \token{</ent>}).\\
    EntMarkersSpans & In addition to the \token{<ent>} and \token{</ent>} tags, global attention is set on the tokens between them, that is, the entities themselves.\\
    EntSpans & We only assign global attention to the entity spans. The \token{<ent>} and \token{</ent>} tokens are replaced by the padding mask token to mask them in inputs and thus do not get either global or local attention. \\
    EntOnly & We additionally mask out all tokens outside the entity spans so they do not get either global or local attention; thus we only pass entities with global attention on them to the decoder. We test this scenario to see how well the summaries can be recovered from only the essential entities plus information collected by \token{<doc-sep>} tokens. \\
    \bottomrule
    \end{tabular}
    \caption{Global attention settings}
    \label{tab:global_attention_setting}
\end{table}

\subsection{Entity marking and global attention}

PRIMERA is based on Longformer-Encoder-Decoder (LED) \citep{beltagy2020longformer}, which uses sparse attention (global attention) in addition to fixed-sized window attention (local attention). Here, we experiment with employing the global attention mechanism
to highlight PICO elements and aggregate them across the documents. Specifically, for the scenario with entity spans in input and target texts, we use the five settings for global attention listed in \Cref{tab:global_attention_setting}. 

\begin{table*}[t!]
\footnotesize
    \centering
    \begin{tabular}{p{0.05cm} p{2cm} p{0.75cm} p{0.75cm} p{0.75cm} c p{0.75cm} p{1.25cm}}
        \toprule
         & &  R-1$\uparrow$ & R-2$\uparrow$ & R-L$\uparrow$ & BERTScore$\uparrow$ & $\Delta$EI$\downarrow$ & $\Delta$EI-$F_1$ $\downarrow$  \\

         \midrule
         
          {\multirow{2}{*}{\rotatebox[origin=c]{90}{\textbf{\ex{Zero}}}}} & Default & 0.215 & 0.032 & 0.132 & 0.834 & 0.580 & 0.321 \\
            & Last 3 & 0.245  & 0.063  & 0.179  & 0.871   & 0.260  & 0.385  \\
         \midrule
         
         {\multirow{6}{*}{\rotatebox[origin=c]{90}{\textbf{\ex{10-shot}}}}} & No entity & 0.229 & 0.037 & 0.147 & 0.857 & 0.269 & 0.328   \\
         
         & DocSep & 0.234 & 0.041 & 0.155 & 0.864 & 0.267 & 0.367 \\
         
         & EntOnly & 0.197 & 0.024 & 0.139 & 0.834 & 0.297 & 0.330 \\
         & EntMarkers & 0.208 & 0.035 & 0.143 & 0.859 & 0.286 & 0.327 \\
         & EntSpans & 0.235 & 0.036 & 0.155 & 0.854  & 0.307 & \textbf{0.295} \\
         & EntMarkersSpans & 0.187 & 0.266 & 0.122 & 0.831 & 0.322 & 0.319 \\
         \midrule
         {\multirow{6}{*}{\rotatebox[origin=c]{90}{\textbf{\ex{100-shot}}}}} & No entity & \textbf{0.259} & 0.052 & 0.171 & 0.864 & 0.302 & 0.376  \\

         & DocSep & 0.251 & 0.048 & 0.164 & 0.862 & 0.339 & 0.452 \\
         
         & EntOnly & 0.237 & 0.038 & 0.157 & 0.851 & 0.308 & 0.389 \\
         & EntMarkers & 0.244 & 0.048 & 0.164 & 0.864 & 0.284 & 0.369 \\
         & EntSpans & \textbf{0.259} & 0.049 & 0.170 & 0.863 & 0.273 & 0.314 \\
         & EntMarkersSpans & 0.251 & 0.048 & 0.166 & 0.863 & 0.301 & 0.315 \\
         \midrule
         {\multirow{6}{*}{\rotatebox[origin=c]{90}{\textbf{\ex{Full}}}}} & No entity & 0.256 & 0.064 & \textbf{0.182} & 0.871 & 0.308 & 0.409 \\
         
         & DocSep & 0.234 & 0.060 & 0.170 & 0.869 & 0.337 & 0.373 \\
         
         & EntOnly & 0.236 & 0.060 & 0.174 & 0.872 & 0.256 & 0.310 \\
         & EntMarkers & 0.244 & \textbf{0.066} & 0.179 & \textbf{0.874} & 0.246 & 0.312 \\
         & EntSpans & 0.237 & 0.061 & 0.174 & \textbf{0.874} & 0.251  & 0.302 \\
         & EntMarkersSpans & 0.230 & 0.059 & 0.168 & 0.873 & \textbf{0.244} & 0.321 \\


         \bottomrule
    \end{tabular}
    \caption{Results of automatic evaluation; $\uparrow$: higher is better, $\downarrow$: lower is better}
    \label{tab:global_attention}
\end{table*}

\subsection{Manipulating inputs}

As dealing with lengthy inputs is a well-known issue for multi-document summarisation, especially in scientific and biomedical domains, we experiment with several settings to control the length of individual input documents: 
\begin{itemize}
\item \textit{Default}: The default PRIMERA setting where LED's token budget of 4096 tokens is distributed evenly across all input documents and they are truncated to the corresponding length.    
\item \textit{Last 3}: In the biomedical domain the most important information appears in conclusions at the end of the paper, so we include only the last three sentences, based on NLTK's sentence tokenizer.\footnote{https://github.com/nltk/nltk}

\end{itemize}

\section{Results}

Tables \ref{tab:global_attention} and \ref{tab:human_eval} report the results of automatic and human evaluation, correspondingly.

\begin{table*}[t!]
\footnotesize
    \centering
    \begin{tabular}{p{0.05cm} p{2.25cm} P{1cm} P{1.25cm} P{1.25cm} P{1.25cm} P{1.25cm} c}
        \toprule
         & &  PICO$\uparrow$ & Direction$\uparrow$ & Factual$\uparrow$ & Contradict.$\downarrow$ & Malformed$\downarrow$ & No evid.$\downarrow$ \\
         \midrule
         {\multirow{2}{*}{\rotatebox[origin=c]{90}{\textbf{\ex{Zero}}}}} & Default & 50 & 15 & 5 & 0 & 0 & 0 \\
         & Last 3 & 50 & 50 & 30 & 0 & 5 & 70\\
         \midrule
         {\multirow{6}{*}{\rotatebox[origin=c]{90}{\textbf{\ex{10-shot}}}}} & No entity & 25 & 45 & 10 & 5 & 30 & 100 \\
         & DocSep & 25 & 50 & 10 & 15 & 20 & 95 \\
         & EntOnly& 10 & 30 & 0 & 10 & 75 & 35\\
         & EntMarkers& 25 & 50 & 15 & 0 & 0 & 70\\
         & Ent
         Spans& 30 & 35 & 5 & 5 & 30 & 65 \\
         & EntMarkersSpans& 20 & 35 & 10 & 5 & 70 & 40\\
         \midrule
         {\multirow{6}{*}{\rotatebox[origin=c]{90}{\textbf{\ex{100-shot}}}}} & No entity & 50 & 50 & 20 & 5& 5 & 60\\
         & DocSep & 50 & 50 & 20& 10& 15& 65\\
         & EntOnly& 45 & 35 & 5 & 5 & 35 & 45 \\
         & EntMarkers& 50 & 45 & 30 & 25 & 25 & 85\\
         & EntSpans& 35 & 40 & 15 & 20 & 10 & 100\\
         & EntMarkersSpans& 60 & 40 & 25 & 0 & 0 & 75\\
         \midrule
         {\multirow{6}{*}{\rotatebox[origin=c]{90}{\textbf{\ex{Full}}}}} & No entity & 50& 60& 35& 10& 10&35\\
         & DocSep & 50 & 50 & 25 & 5 & 10 & 65 \\
         & EntOnly& 30 & 40 & 20 & 0 & 5 & 85\\
         & EntMarkers & 35 & 40 & 20 & 10 & 0 & 90\\
         & EntSpans& 55& 40& 25 & 5 & 5 & 90\\
         & EntMarkersSpans& 50 & 40 & 25 & 5 & 0 & 100\\
         \bottomrule
    \end{tabular}
    \caption{Results of human evaluation; $\uparrow$: higher is better, $\downarrow$: lower is better. \textbf{\ex{Zero}} denotes the zero-shot setting.}
    \label{tab:human_eval}
\end{table*}

\subsection{Models with and without global attention on entities}

Though we do not see major improvements in ROUGE scores between the model without PICO entity marking (\textit{No entity}) and the models with global attention on PICO entities (with the exception of \textit{EntMarkers} and \textit{EntSpans}) and even observe some decrease in factuality scores, on closer inspection the summaries generated by those systems prove to be qualitatively different. In particular, the \textit{No entity} model is more extractive and more extensively copies the input studies, while the results of models with global attention on entities are more abstractive. For example, for review CD005963 (\Cref{tab:examples} in Appendix \ref{gen_examples}), the \textit{No entity} model copies the term \textit{Mental Health Act} often mentioned in source documents but absent in target conclusions, while the other models do not.

\begin{table*}[t]
\footnotesize
    \centering
    \begin{tabular}{p{0.05cm} p{2cm} p{0.75cm} p{0.75cm} p{0.75cm} c p{0.75cm} P{1.35cm}}
         & &  R-1$\uparrow$ & R-2$\uparrow$ & R-L$\uparrow$ & BERTScore$\uparrow$ & $\Delta$EI$\downarrow$ & $\Delta$EI-$F_1$ $\downarrow$  \\
         \midrule
          {\multirow{4}{*}{\rotatebox[origin=c]{90}{\textbf{\ex{Default}}}}} & Zero-shot & 0.215 & 0.032 & 0.132 & 0.834 & 0.580 & \textbf{0.321} \\
         & 10-shot & 0.229 & 0.037 & 0.147 & 0.857 & 0.269 & 0.328  \\
         & 100-shot & \textbf{0.259} & 0.052 & 0.171 & 0.864 & 0.302 & 0.376  \\
         & Full & 0.256 & \textbf{0.064} & \textbf{0.182} & 0.871 & 0.308 & 0.409  \\
         \midrule
          {\multirow{4}{*}{\rotatebox[origin=c]{90}{\textbf{\ex{Last 3}}}}} & Zero-shot & 0.245  & 0.063  & 0.179  & \textbf{0.871}   & \textbf{0.260}  & 0.385  \\
         & 10-shot & 0.211  & 0.030  & 0.135  & 0.853  & 0.289  & 0.342  \\
         & 100-shot & 0.250  & 0.046  & 0.164  & 0.862  & 0.341  & 0.424  \\
         & Full & 0.239  & 0.061  & 0.171  & 0.870  & 0.279  & 0.382   \\

    \end{tabular}
    \caption{Results of automatic evaluation; $\uparrow$: higher is better, $\downarrow$: lower is better}
    \label{tab:manip_input}
\end{table*}

 \Cref{tab:abstractive} in Appendix \ref{lex_overlap} shows how the overlap with source documents decreases when the entity marking with global attention is used, thus making the summaries more abstractive. This, however comes at a cost: we notice that the models with additional global attention produce remarkably more \textit{no evidence} summaries, and in the fully fine-tuned scenario the number of such summaries grows with the number of tokens on which we place global attention. This is consistent with the results of another model which extensively uses global attention \citep{deyoung-etal-2021-ms} which also produces a large number of \textit{no evidence} summaries  \citep{otmakhova-etal-2022-patient}. Another behaviour of models with extra global attention observed both in \citet{deyoung-etal-2021-ms} and here is that they generate sequences which are representative of biomedical text style. For example, in addition to conclusions, the summaries generated by such models contain generic sentences such as \textit{There is a need for more studies of high methodological quality}. Thus we hypothesise that tokens with global attention tend to accumulate and reproduce information common to a large number of documents in the training set rather than information shared by a particular set of input documents.
 Finally, though we expected the \textit{EntOnly} model, which only uses only PIO entities as inputs and thus loses information about the relations between them, to perform much worse than the other models, it is very similar to them both in automatic metrics and \textit{Direction} scores. We maintain that it shows that even if the models are able to attend to all tokens, they only reproduce PIO entities and are not able to consistently capture the relationships between them.

\subsection{Zero-shot vs.\ few-shot vs.\ fully fine-tuned models}

We notice that in terms of automatic metrics, zero-shot models are comparable to fine-tuned ones or even outperform them; however they perform substantially worse in terms of factuality, especially for the direction. We find that in zero-shot scenarios, PRIMERA copies spans of text from one or several of the input documents, focusing mostly on their beginnings, rather than aggregates information across documents. Thus it outputs either conclusions copied from a single document, or, more often, makes no claims at all by reporting the objectives of the review or its setup.

Another interesting finding is that the ROUGE scores tend to be the highest in the 100-shot scenario and go down for the fully fine-tuned models. We maintain that in 10-shot scenarios the models are still unable to correctly capture and reproduce important entities (which is also reflected in their low accuracy in terms of PICO), while in the fully fine-tuned models, there is a tendency to generate broader and generic entities, for example \textit{metal-protein attenuation compounds} instead of \textit{PBT1/PBT2} in the target summary.

Not surprisingly, the number of malformed generations decreases with increasing the number of training samples: the majority of summaries produced by \textit{EntOnly} and \textit{EntMarkersSpans} after 10 shots are malformed, but even 100-shot training significantly reduces this amount. On the other hand, it is surprising to see that the more the models are fine-tuned the more \textit{no evidence} statements they produce, with some models generating only such summaries in fully fine-tuned scenario.

Lastly, we find that the 100-shot \textit{EntMarkers} model is similar in terms of factuality to the fully fine-tuned model without entity marking (\textit{No entity}). This is an encouraging result as high-quality multi-document summarisation data is scarce in biomedical domain, so few-shot learning is a practically important direction to explore.

\subsection{\textit{Default} vs.\ \textit{Last3}}

For few-shot and fine-tuned models we find no major improvements in quality when restricting the inputs to the last three sentences only (\Cref{tab:manip_input}). This shows that after fine-tuning PRIMERA is able to detect most useful spans without relying on their explicit marking. On the other hand, for the zero-shot scenario, where the model tends to copy from the beginning of input documents, the quality dramatically improves when we force it to extract only from a more informative span at the end of documents. Interestingly, such an easy manipulation of inputs allows to achieve results comparable to the best 100-shot and fully fine-tuned models without any training on the in-domain dataset. Again, this is a promising direction for research considering the scarcity of high-quality data.

\section{Conclusion}
We tackle the problem of biomedical multi-document summarisation by incorporating PICO information into a strong summarisation model, and using global attention to enhance the representation of this information. Through automatic and human evaluations on an extensive set of experiments, we find that adding global attention to PICO spans would help in (1) generating more abstractive summaries, and (2) improving summarization quality in few-shot settings, which is especially important in the biomedical domain.


\section*{Acknowledgements}

The authors would like to thank the anonymous reviewers for their comprehensive and constructive reviews. This research was undertaken using the LIEF HPC-GPGPU Facility hosted at the University of Melbourne. This Facility was established with the assistance of LIEF Grant LE170100200. This research was conducted by the Australian Research Council Training Centre in Cognitive Computing for Medical Technologies (project number ICI70200030) and funded by the Australian Government.

\bibliography{anthology,custom}
\bibliographystyle{acl_natbib}


\appendix

\section{Dataset statistics}

\label{dataset_stats}

Table~\ref{tab:cochrane_statistic} reports some basic statistics of the Cochrane dataset used in this challenge. 
The Average number of PICO spans in the summary and the input documents (Avg. \# PICO spans) are obtained using the trained PICO tagger.
Note that target summaries for the test set are not provided.

\begin{table}[!htbp]
\footnotesize
    \centering
    \begin{tabular}{p{2.5cm} p{1cm} p{1cm} p{1cm}}
    \toprule
        & Train & Valid. & Test  \\
    \midrule
    \# samples & 3752 & 470 & 470 \\
    Avg. input length & 2417 & 2389 & 2677 \\
    Avg. summary length & 68 & 70 & n/a \\
    Avg. \# PICO spans in input  & 213 & 209 & 236 \\
    Avg. \# PICO spans in summary  & 4 & 4 & n/a \\
    
    \bottomrule
    \end{tabular}
    \caption{Cochrane dataset statistics.}
    \label{tab:cochrane_statistic}
\end{table}



\section{Examples of malformed, contradictory and non-evidential summaries}
\label{sec:examples}

To clarify the criteria we used for evaluation, Table \ref{tab:errors} lists some examples of contradictory, malformed and non-evidential summaries. Malformed summaries are ones containing repetitions, incomplete text or corrupted tokens. The spans of text corresponding to errors are in \textbf{bold}.

\begin{table*}[t]
\footnotesize
    \centering
    \begin{tabular}{p{2cm} p{12cm}}
    \toprule
    Error & Summary \\
    \midrule

       \multirow{2}{*}{Contradiction}     & \textit{There is \textbf{insufficient evidence} to support the use of edaravone as a therapy for acute ischemic stroke. However, it may be useful for treating other types of ischemic stroke. The current review \textbf{provides a rationale basis} for the use of edaravone as a therapy for acute ischemic stroke.}\\
      
      & \textit{\textbf{In the absence of evidence to support} the use of PBT2 in patients with severe Alzheimer's disease, clinicians and patients \textbf{should recommend} the continued use of PBT2.}\\
    \midrule
   \multirow{3}{*}{Malformed}       &  \textit{There is inadequate evidence to evaluate the effect of percutaneous endoscopic gastrostomy on the incidence of percutaneous wound infections. The current evidence base is limited due to the differing methodologies employed in the trials. \textbf{The current evidence base is limited due to the differing methodologies employed in the trials. The current evidence base is limited due to the differing methodologies employed in the trials...}} \\
   & \textit{We found no clear evidence to support the use \textbf{of}}\\
   & \textit{There is limited evidence to suggest that the use of \textbf{apleuapleuapleuapleuapleuapleua...}}\\
    \midrule
    No evidence &  \textit{There is insufficient evidence to support the use of metal-protein-attenuating compounds for the treatment of AD. Further trials are needed.}\\

    \bottomrule
    
\end{tabular}
\caption{Examples of contradictory, malformed and non-evidential summaries}
\label{tab:errors}

\end{table*}

\section{Examples of generated summaries}
\label{gen_examples}

\Cref{tab:examples} shows the examples of summaries generated for input documents for review CD005963.

\begin{table}[t]
\footnotesize
    \centering
    \begin{tabular}{p{2cm} p{5cm}}
    \toprule
    Setting & Summary \\
    \midrule

      No entity     & \textit{... the results suggest that advance directives may be beneficial in reducing the number of people admitted to hospital under the Mental Health Act.}                                       \\
    
    DocSep       &  \textit{There is insufficient evidence to support or refute the use of advance directives for people with mental illnesses. } \\
    
    EntMarkers &  \textit{There is insufficient evidence to support or refute the use of advance directives for people with severe mental illness.}\\
    
    EntMarkersSpans &  \textit{There is insufficient evidence to support the use of advance directives for people with severe mental illness.}\\
    
    EntSpans & \textit{There is insufficient evidence to support the use of advance directives for people with mental illness.}\\
    
    EntOnly & \textit{There is insufficient evidence to support the use of advance directives for people with severe mental illness.}\\
    
    \bottomrule
    
\end{tabular}
\caption{Examples of generated summaries}
\label{tab:examples}

\end{table}

\section{Lexical overlap with the input documents}
\label{lex_overlap}

\Cref{tab:abstractive} shows the amount of lexical overlap with the source documents in terms of ROUGE scores. The lower the score is, the less is copied from the source and the more abstractive the summary is.

\begin{table}[h]
\footnotesize
    \centering
    \begin{tabular}{p{0.05cm} p{2cm} p{0.75cm} p{0.75cm} p{0.75cm}}
         & &  R-1$\downarrow$ & R-2$\downarrow$ & R-L$\downarrow$\\

                  \midrule
         
         {\multirow{6}{*}{\rotatebox[origin=c]{90}{\textbf{\ex{Full}}}}} & No entity & 0.052 & 0.022 & 0.040 \\
         
         & DocSep & 0.042 & 0.019 & 0.034 \\
         
         & EntOnly & 0.043 & 0.021 & 0.036 \\
         & EntMarkers & 0.042 & 0.018 &  0.033 \\
         & EntSpans & 0.040 & 0.017 & 0.032 \\
         & EntMarkersSpans & 0.037 & 0.016 & 0.030 \\

         \bottomrule
    \end{tabular}
    \caption{Token overlap with the source as a measure of extractiveness; lower = more abstractive}
    \label{tab:abstractive}
\end{table}




\end{document}